\pdfoutput=1

\documentclass[11pt]{article}

\usepackage[final]{acl}

\usepackage{times}
\usepackage{latexsym}

\usepackage[T1]{fontenc}

\usepackage[utf8]{inputenc}

\usepackage{microtype}

\usepackage{inconsolata}

\usepackage{graphicx}

\usepackage{makecell}
\usepackage{caption}
\usepackage{subcaption}
\usepackage{multirow}
\usepackage{menukeys}
\usepackage{multicol}

\title{Leveraging Large Language Models\\ to Geolocate Linguistic Variations\\ in Social Media Posts}

\author{%
Davide Savarro\textsuperscript{{\normalfont 1},\thanks{Equal contribution.}}\And
Davide Zago\textsuperscript{{\normalfont 1}, *}\And
Stefano Zoia\textsuperscript{{\normalfont 1}, *}\AND
{\normalfont \small \textsuperscript{1}Department of Computer Science, University of Turin,}\\
{\small Corso Svizzera 185, 10149 Torino, Italy}\\
{\small \texttt{\{davide.savarro, davide.zago, stefano.zoia\}@unito.it}}}

\begin{document}
\maketitle

\begin{abstract}
Geolocalization of social media content is the task of determining the geographical location of a user based on textual data, that may show linguistic variations and informal language. In this project, we address the GeoLingIt challenge of geolocalizing tweets written in Italian by leveraging large language models (LLMs). GeoLingIt requires the prediction of both the region and the precise coordinates of the tweet. Our approach involves fine-tuning pre-trained LLMs to simultaneously predict these geolocalization aspects. By integrating innovative methodologies, we enhance the models' ability to understand the nuances of Italian social media text to improve the state-of-the-art in this domain. This work is conducted as part of the Large Language Models course at the Bertinoro International Spring School 2024. We make our code publicly available on GitHub \footnote{\url{https://github.com/dawoz/geolingit-biss2024}}.
\end{abstract}

\section{Introduction}

\textit{GeoLingIt} is the first shared task focused on geolocating linguistic variation in Italy using social media posts with non-standard Italian language. Part of the EVALITA evaluation campaign, it aims to advance natural language processing (NLP) for non-standard Italian and provide sociolinguistic insights through quantitative analysis.

Social media offers a unique opportunity to study informal language use across sociolinguistic dimensions, particularly diatopic variation (variation across geographic space). Italy's linguistic diversity includes numerous local languages, dialects, and regional varieties of Standard Italian. Online, Italians often use local language elements to signal social identities.

GeoLingIt seeks to understand linguistic variation in Italy by developing methods to predict the locations of Twitter posts based solely on linguistic content. Unlike other geolocation tasks, it filters posts for non-standard Italian, focusing on linguistic rather than lexical variations. Variations in GeoLingIt data may include local words, code-switching, or entire posts in a local language or dialect.

GeoLingIt includes two subtasks. Coarse-grained geolocation, named subtask A, consists in the classification of the region of provenance of a given text. Fine-grained geolocation, named subtask B, is a double regression task, in which the algorithm must predict the coordinates of the given social media post.

Our approach involves fine-tuning pre-trained large language models (LLMs) to solve both subtasks in a single generation step. Our methodology is inspired by ExtremITA \cite{hromei2023extremita}, where a multi-task approach was used to train LLMs to solve all EVALITA tasks. We fine-tune and compare three decoder-only LLMs on GeoLingIt and analyze the obtained results.

\section{Data and models}

In this section we describe the GeoLingIt dataset and we outline the characteristics of the three LLMs used in our experiments. We also discuss the pre-processing steps and the methodologies employed for fine-tuning these models.

\subsection{Dataset description}
\label{subsec:DataMod}

The GeoLingIt dataset contains 15039 samples and it is divided in a specific train-evaluation-test split of 13669 (90.09\%), 552 (3.67\%) and 818 (5.44\%) samples, to enable a fair comparison between the different approaches. Furthermore, the dataset is divided into 2 subsets that correspond respectively to subtask A and B. Since our approach tackles both tasks simultaneously, we join the two portions into a new merged dataset and we use it during training. \autoref{tab:dataset} shows some of the samples of the resulting set of data.
 
\begin{table*}
    \footnotesize
    \centering
    \begin{tabular}{l|l|l|l|l}
        Id & Text & Region & Latitude & Longitude \\
         \hline
         \texttt{280} & \texttt{[USER] A suma bin ciapa'! meglio alleggerire un attimo} & \texttt{Piemonte} & \texttt{45.0729} & \texttt{7.6758}\\
         \texttt{286} & \texttt{[USER] Ce ripigliamm tutt chell ch è o nuost} & \texttt{Campania} & \texttt{40.8541} & \texttt{14.2435}  \\
         \texttt{500} & \texttt{[USER] [USER] Sta bon, vecio!} & \texttt{Veneto} & \texttt{46.1572} & \texttt{12.2865}
    \end{tabular}
    \caption{Some samples of the GeoLingIt dataset. The column \textit{Text} includes the content of the post. This can contain urls, user tags and images that are replaced with placeholders. The region of provenance is annotated under the column \textit{Region}, and the coordinates under the columns \textit{Latitude} and \textit{Longitude}.}
    \label{tab:dataset}
\end{table*}

\begin{figure}
    \centering
    \includegraphics[width=\linewidth]{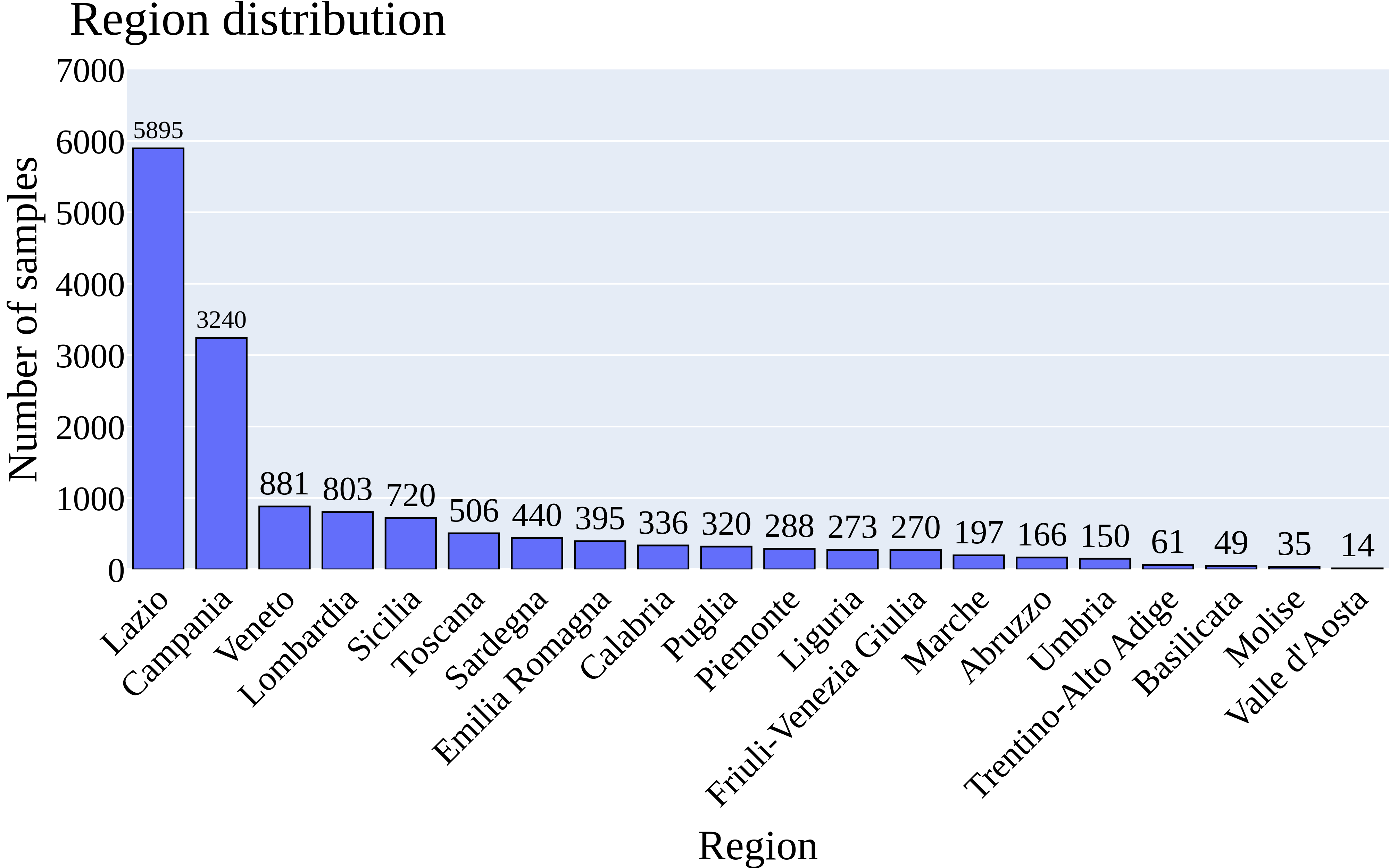}
    \caption{Bar chart representing the number of posts for each label (region of provenance). The labels (x-axis) are sorted by descreasing frequency (y-axis).}
    \label{fig:region_distribution}
\end{figure}

Regarding the prediction of the region of provenance (subtask A), the dataset contains 20 labels, which correspond to all the regions of Italy. The distribution of such labels is very nonuniform, as shown in \autoref{fig:region_distribution}: 39.20\% of the samples are labeled with the \textit{Lazio} region, and 21.54\% with the \textit{Campania} region, the two subsets alone constituting the 60.74\% of the dataset.

\begin{figure*}
     \centering
     \begin{subfigure}[t]{.47\linewidth}     
        \centering
        \includegraphics[width=1\linewidth]{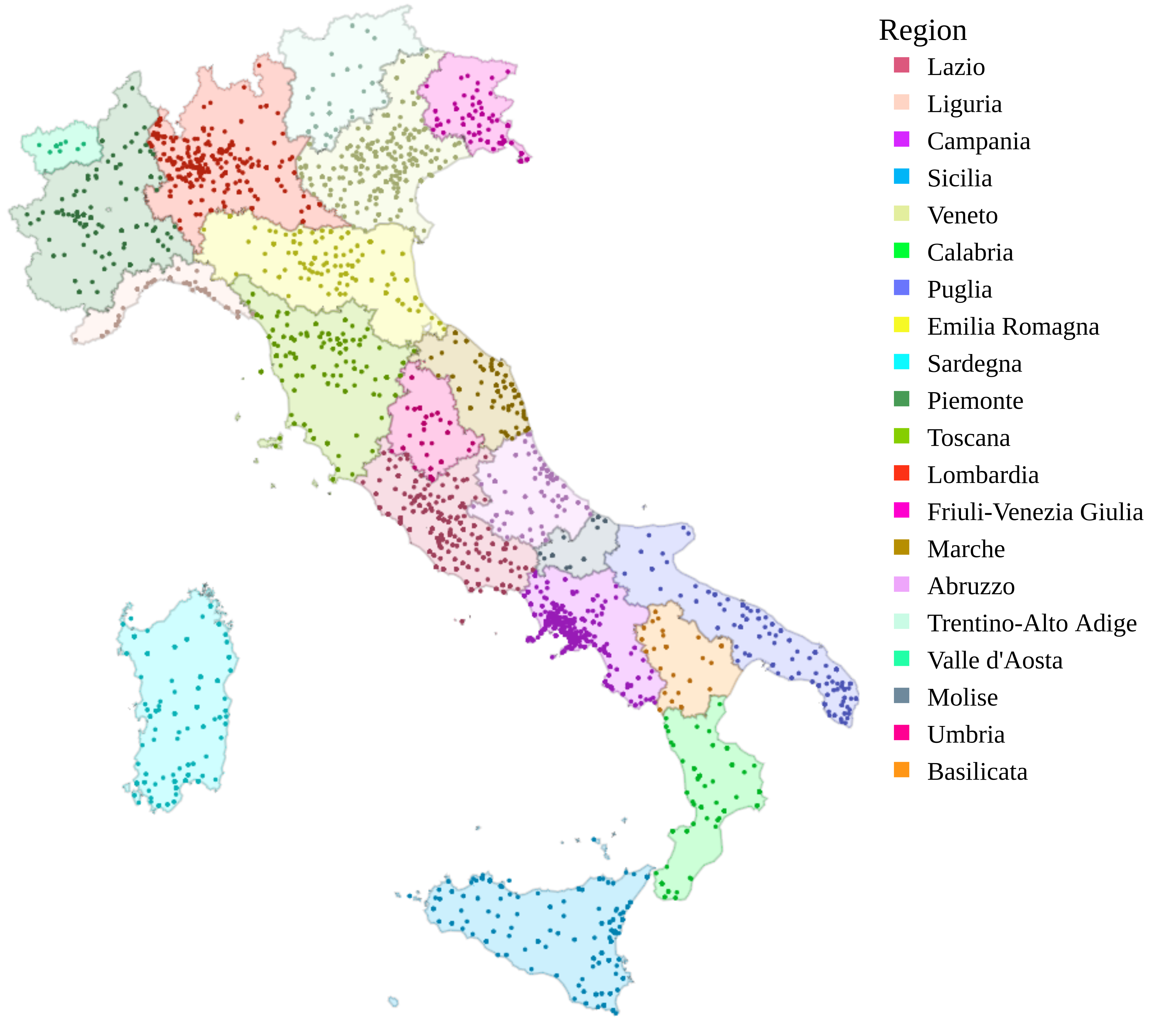}
        \caption{Exact locations of the posts in the dataset. For clarity, the different regions of Italy are displayed with different colors.}
        \label{fig:coordinate_distribution}
     \end{subfigure}
     \hfill
     \begin{subfigure}[t]{.47\linewidth}
        \centering
        \includegraphics[width=\linewidth]{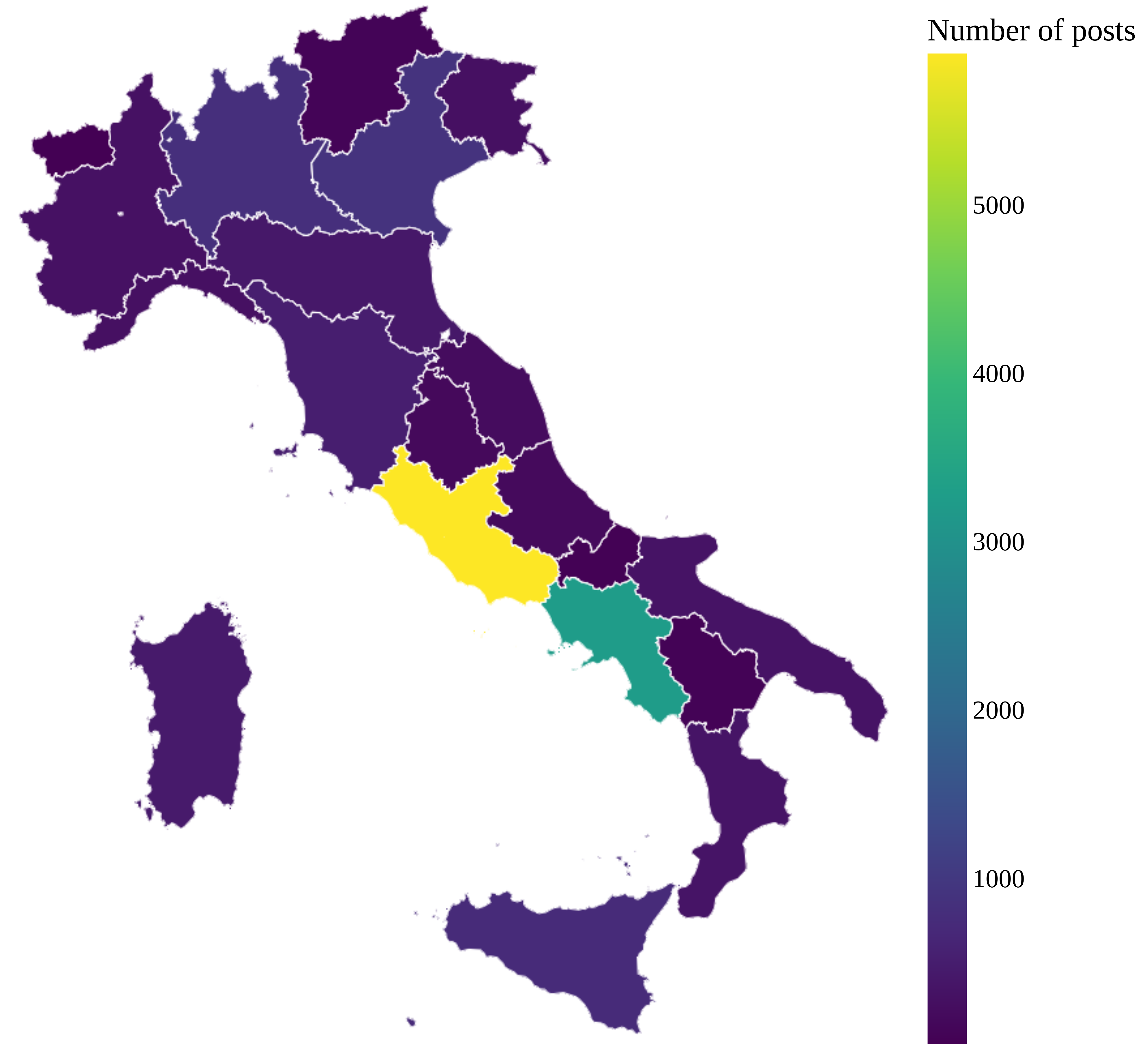}
        \caption{Density of the locations of social media posts. Lower densities are coloured with a dark colour (purple) and higher densities are with brighter colors (yellow).}
        \label{fig:number_of_posts_per_region}
     \end{subfigure}
        \caption{Geographical distribution of the social media posts in the GeoLingIt dataset.}
        \label{fig:coordinate_distribution_and_number_of_posts_per_region}
\end{figure*}

The locations of the social media posts are displayed in the map in \autoref{fig:coordinate_distribution}, in which different regions are filled with different colors. Each point corresponds to a specific latitude-longitude pair associated to a sample. From the figure it is evident that the frequency of posts increase in highly populated areas, like in the regional capitals. \autoref{fig:number_of_posts_per_region} shows in the geolocation map the density of posts in each region. As a rule of thumb, regions with higher number of posts are regions with higher population, but not all densely populated regions are proportionally represented in the dataset (e.g. \textit{Lombardia}, which has the highest population).

\subsection{Models description}

In recent years pre-trained LLMs have been effectively used to solve varied problems in natural language processing. In our setting, we fine-tune and compare three different pre-trained LLMs on the Italian language: Camoscio-7B \cite{santilli2023camoscio}, ANITA-8B \cite{polignano2024llamantino_anita} and Minerva-3B \cite{orlando2024minerva}.

\textbf{Camoscio} is an Italian instruction-tuned 7 billion parameters model based on LLaMA \cite{touvron2023llama}. The training of the model follows the one of \cite{stanford_alpaca} with Low-Rank Adaptation (LoRA) \cite{hu2021lora}. We use a slightly modified version of that model implemented by the ExtremITA \cite{hromei2023extremita} team.

\textbf{ANITA} is a 8 billion parameter instruction-tuning of the LLaMAntino family \cite{basile2023llamantino}. This model was obtained by fine-tuning LLaMA 3 \cite{llama3modelcard} with Direct Preference Optimization (DPO) \cite{rafailov2024dpo}. ANITA aims to be a multilingual model to be used for further fine-tunings on Italian language tasks.

\textbf{Minerva} is the first family of LLMs pre-trained from scratch on the Italian language. This set includes three model sizes of 350 millions, 1 billion and 3 billions parameters respectively. We include Minerva in our study to test the capability of a pre-trained LLM on Italian, and we use the largest (3 billion) version to obtain a comparison that is as fair as possible with the other larger models.

\section{Experiments and result analysis}
We will now briefly present the hardware used, the experiments performed and we will analyze the metrics and the results we obtained. 

\subsection{Hardware}
\label{subsec:HWLim}

All our experiments are run on a single machine equipped with a \textit{Tesla T4} GPU with 16 GB of VRAM. Due to the limited amount of compute power, we had to make some compromises when choosing how to train and test our models (see subsection \ref{subsec:ExpSet}).  

\subsection{Experiment settings}
\label{subsec:ExpSet}

All the previously mentioned models have been fine-tuned and tested on the GeoLingIt dataset following the ExtremITA instruction encoding: 

\phantom{.}
\newline
\noindent $\langle instruction \rangle \langle post\_content \rangle$
\keys{\return}

\texttt{[regione]} $\langle region \rangle$ \texttt{[geo]} $\langle lat \rangle \langle long \rangle$ 
\newline

\noindent In this way, both subtasks A and B have been solved with a single model fine-tuning with the language modeling objective. 

The upper panel of \autoref{tab:hyp} shows the hyperparameters we used for fine-tuning across all the experiments except for Minerva, for which we used a larger batch and minibatch size. Due to our limited disposal of GPU memory of 16 GB, we used gradient accumulation to simulate larger batch sizes and stabilize training.

To further reduce the memory footprint we used 4-bit quantized versions of the three considered models, and we used LoRA to reduce the number of training parameters and therefore the training time. Although smaller models such as Minerva (3 bilion parameters) would have benefited from a more conservative LoRA configuration (e.g. higher rank values or greater bit precision), those same settings would not have been viable for Camoscio and ANITA with our computational resources. For these reasons, we kept the same configuration for all the tested models.  The lower panel of \autoref{tab:hyp} sums up the parameters we used for LoRA with quantization.

It is worth noting that we performed 10 epochs of training for each of the models analyzed in order to compensate for the aggressive LoRA settings. In this way we were able to still get some noteworthy results when comparing them with the ones presented in \cite{hromei2023extremita}.    

\begin{table}
\begin{center}
\begin{tabular}{ |c|c|c| } 
\hline
& \textbf{Parameter} & \textbf{Value} \\
\hline
\parbox[t]{7px}{\multirow{5}{*}{\rotatebox[origin=c]{90}{Fine-tuning}}} & N Epochs & \(10\) \\ 
& Batch Size & \(32\) (\(64^*\) ) \\ 
& M. Batch Size & \(8\) (\(32^*\)) \\
& Learning Rate & \(3 \cdot 10^{-4}\) \\
& Warmup Ratio & \(0.1\) \\
\hline
\parbox[t]{7px}{\multirow{4}{*}{\rotatebox[origin=c]{90}{QLoRA}}} & R & \(8\) \\ 
& Alpha & \(16\) \\ 
& Dropout & \(0.05\) \\
& N. Bits & \(4\) \\
\hline
\end{tabular}
\end{center}
\caption{Hyperparameters used for fine-tuning ($^*$ parameters used for the Minerva model). The parameters used for quantized LoRA are shown in the bottom part of the table.}
\label{tab:hyp}
\end{table}

\subsection{Metrics and result analysis}
We choose to use the same metrics that were adopted for the EVALITA 2023 campaign to evaluate our models and enable a fair comparison with the other competitors: macro (unweighted) F1-score for subtask A, and average distance error in km for subtask B.

The average distance error is a precise and intuitive metric when comparing the regressive capabilities of the models in predicting the right coordinates. On the other hand, we argue that the macro F1-score is not the best metric for evaluating region classification capabilities in our setting. Given the fact that the GeoLingIt dataset is highly unbalanced (see subsection \ref{subsec:DataMod} for details), we expect that most represented classes would show lower prediction error, and, on the contrary, samples of less represented classes would be more often misclassified. We believe that the F1-score weighted by class cardinality (also known as "micro") is a more appropriate choice of metric, and envision additional experiments with this considerations.

Analyzing the results reported in \autoref{tab:res} it is interesting to see how a 1 year evolution of LLM research actually influences the performances. We can assert the performance of Minerva is comparable with Camoscio even though Minerva is less than half the size of Camoscio. Moreover, ANITA produced very strong results, almost matching the ones obtained by the top performing models of the EVALITA 2023 campaign. 

\begin{table}
\begin{center}
\begin{tabular}{ |l|c|c| } 
\hline
\textbf{Model} & \textbf{F1-score} (macro) & \textbf{Avg Km}\\
\hline
Camoscio-7B & \(0.4935\) & \(124.35\) \\ 
ANITA-8B & \(0.5411\) & \(103.06\) \\ 
MINERVA-3B & \(0.4704\) & \(125.35\) \\ 
\hline 
W\_EVALITA23 & \(0.6630\) & \(97.74\) \\
\hline
\end{tabular}
\end{center}
\caption{Experiment results, compared with the best-performing models in EVALITA 2023 \cite{Ramponi2023GeoLingItAE}}
\label{tab:res}
\end{table}

\section{Error analysis}

For subtask A (region classification), the models showed varying performance in classifying the region of provenance, particularly struggling with less represented regions in the dataset (e.g. \textit{Trentino-Alto Adige}). Due to space reasons we include the confusion matrices of the test set classification in the \nameref{sec:appendix}. Figures \ref{fig:confusion_matrix_camoscio}, \ref{fig:confusion_matrix_anita} and \ref{fig:confusion_matrix_minerva} reveal that the most frequent errors occurred in predicting regions with fewer samples. As expected, the models tend to favor regions with higher representation, such as \textit{Lazio} and \textit{Campania}, leading to biased predictions. It is still interesting to see that often misclassifications happen between nearby regions (e.g. \textit{Umbria} and \textit{Lazio} or \textit{Piemonte} and \textit{Lombardia}) highlighting linguistic similarities that are difficult for the LLM to distinguish. 

In subtask B (coordinate regression), predicting precise coordinates posed a challenge due to the fine-grained nature of the task. Figures \ref{fig:regression_error_llama_mean}, \ref{fig:regression_error_anita_mean} and \ref{fig:regression_error_minerva_mean} show the average distance errors in Italian provinces. It is evident from the figures that while the models could approximate the general area, pinpointing exact locations was difficult. ANITA-8B outperformed the others, likely due to its larger parameter size and advanced training techniques.

Observing the sum of distance errors evidences how much the models predicted distant values despite the number of samples, as shown in figures \ref{fig:regression_error_llama_sum}, \ref{fig:regression_error_anita_sum} and \ref{fig:regression_error_minerva_sum}. From these figures we can see that non negligible amounts of error also come from frequently represented areas (see \autoref{fig:number_of_posts_per_region}).

\section{Conclusions}

This project tackled the challenge of geolocating social media posts written in non-standard Italian using large language models (LLMs). By fine-tuning pre-trained LLMs, we aimed to predict both the region and precise coordinates of tweets, addressing the GeoLingIt subtasks simultaneously. Our experiments with Camoscio-7B, ANITA-8B, and Minerva-3B demonstrated encouraging resulsts in geolocation accuracy. Despite the dataset's non-uniform distribution posing challenges, our experiments showed that recent advancements in LLMs significantly enhance geolocation performance, with the ANITA-8B model achieving the best results among the tested models.

Future work could explore advanced pre-processing techniques (e.g. including additional information in the prompt, like the coordinate centroid of a given region) and different models to further improve geolocation accuracy. Additionally, class imbalance issues could be addressed with data augmentation or re-sampling techniques.

Overall, this project contributes to the research on using LLMs for sociolinguistic analysis and geolocation tasks, offering a foundation for developing more accurate and sophisticated models.

\section*{Acknowledgements}

We thank the Department of Computer Science of the University of Turin for providing the access to HPC4AI with the necessary computational resources for running the experiments of this work.
        
\bibliography{main}

\clearpage

\appendix
\onecolumn

\section*{Appendix}
\label{sec:appendix}

\vspace{30px}

\begin{figure*}[h]
    \centering
    \begin{subfigure}[c]{.48\linewidth}     
        \centering
        \includegraphics[width=\linewidth]{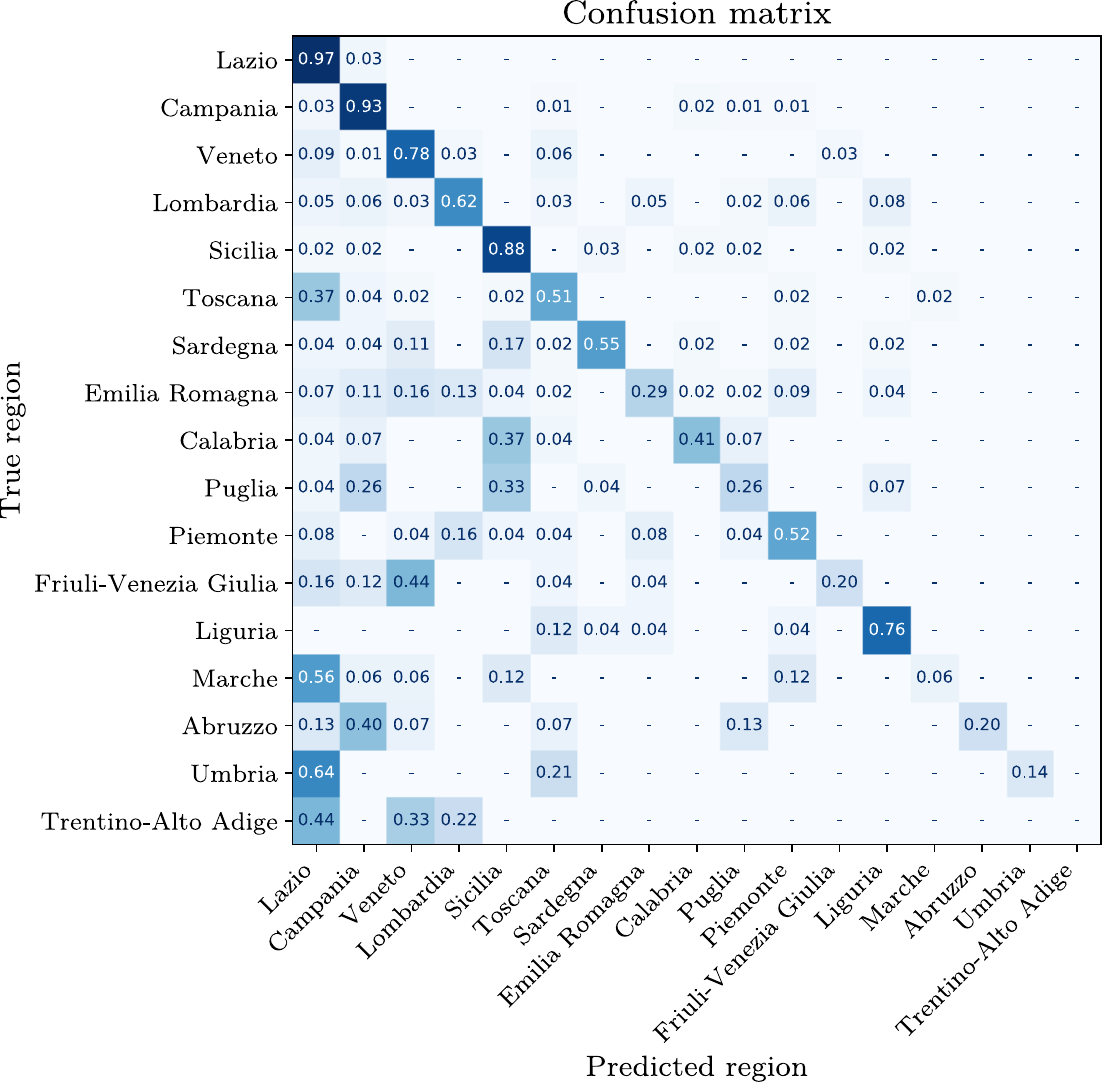}

        \vspace{5px}
        
        \caption{Confusion matrix for the classification with Camoscio.}
        \label{fig:confusion_matrix_camoscio}
    \end{subfigure}\hfill     
    \begin{subfigure}[c]{.48\linewidth}
       \centering
       \includegraphics[width=\linewidth]{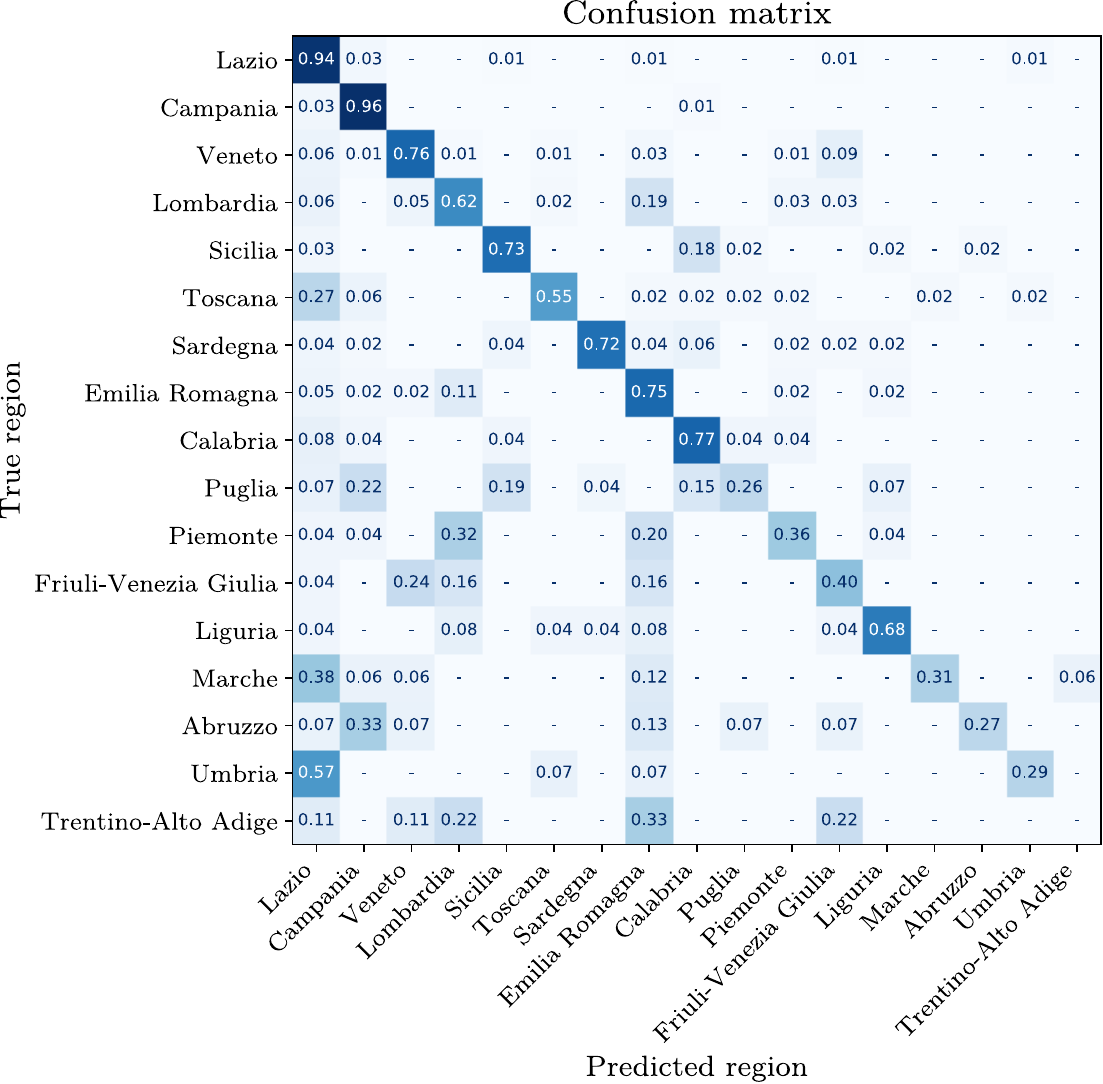} 
        
        \vspace{5px}
        
        \caption{Confusion matrix for the classification with ANITA.}
        \label{fig:confusion_matrix_anita}
    \end{subfigure} 
    
    \vspace{20px}
    
    \begin{minipage}[c]{.48\linewidth}
        \begin{subfigure}[c]{\linewidth}
           \centering
           \includegraphics[width=\linewidth]{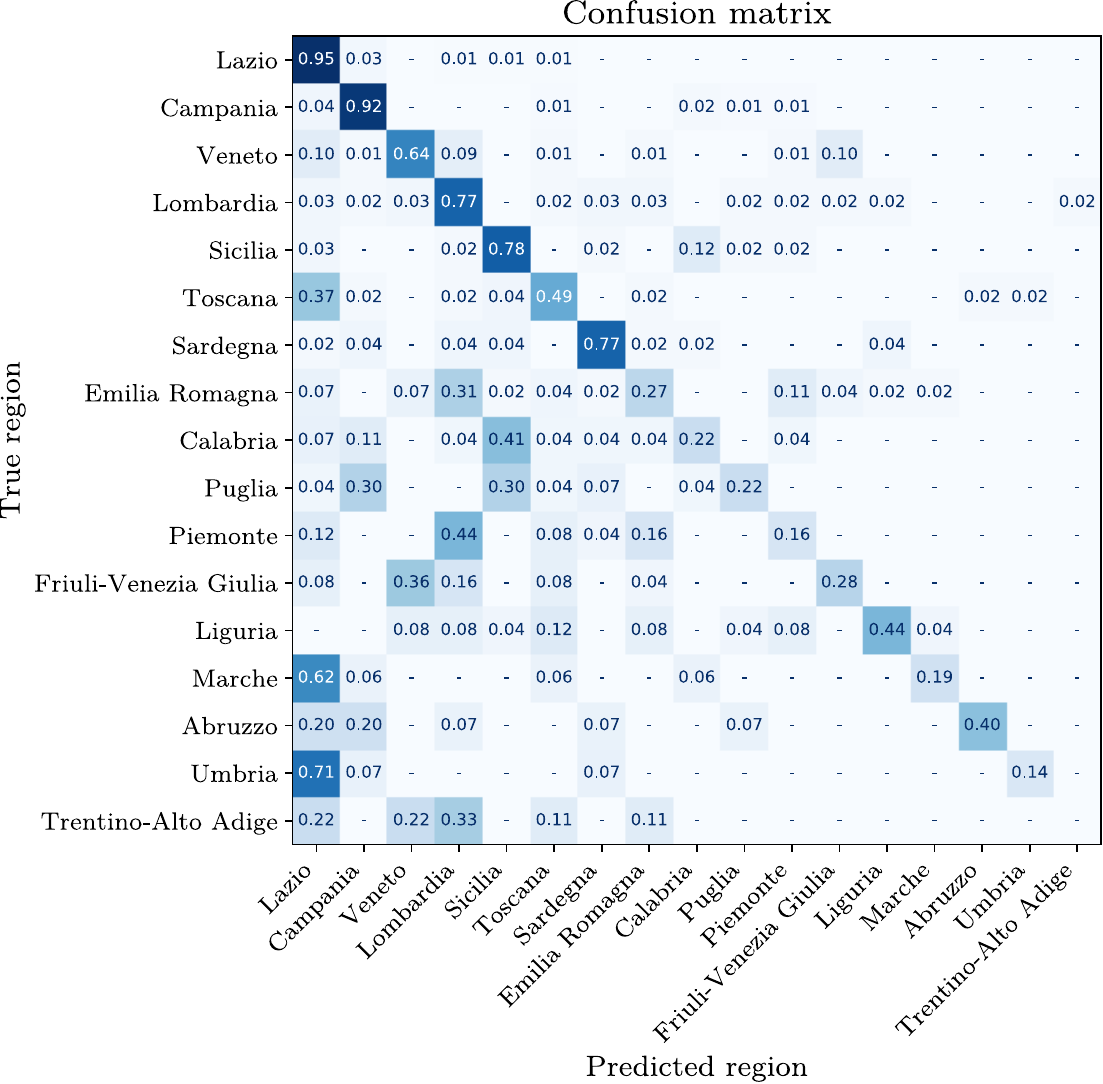}    

            \vspace{5px}
           
            \caption{Confusion matrix for the classification with Minerva.}
            \label{fig:confusion_matrix_minerva}
        \end{subfigure}
    \end{minipage}\hfill
    \begin{minipage}[c]{.43\linewidth}
         \caption{Confusion matrices for the classification of the samples in the test set for all tested models: Camoscio (\ref{fig:confusion_matrix_camoscio}), ANITA (\ref{fig:confusion_matrix_anita}) and Minerva (\ref{fig:confusion_matrix_minerva}). The classes on the x and y axis include only the classes present in the test set, which are a subset of Italian regions. The numbers in each cell $(c_{pred},c_{true})$ correspond to the frequency of samples with class $c_{true}$ classified as $c_{pred}$ and normalized by the total number of samples of the true class (row). Cells containing "-" mean zero frequency of classified samples. Darker colors highlight higher frequencies, and a darker main diagonal on the matrix implies strong classification performance.}
     \end{minipage}
    \label{fig:confusion_matrix}
\end{figure*}    

\clearpage

\begin{figure*}
     \centering
     \begin{subfigure}[t]{.44\linewidth}     
        \centering
        \includegraphics[width=1\linewidth]{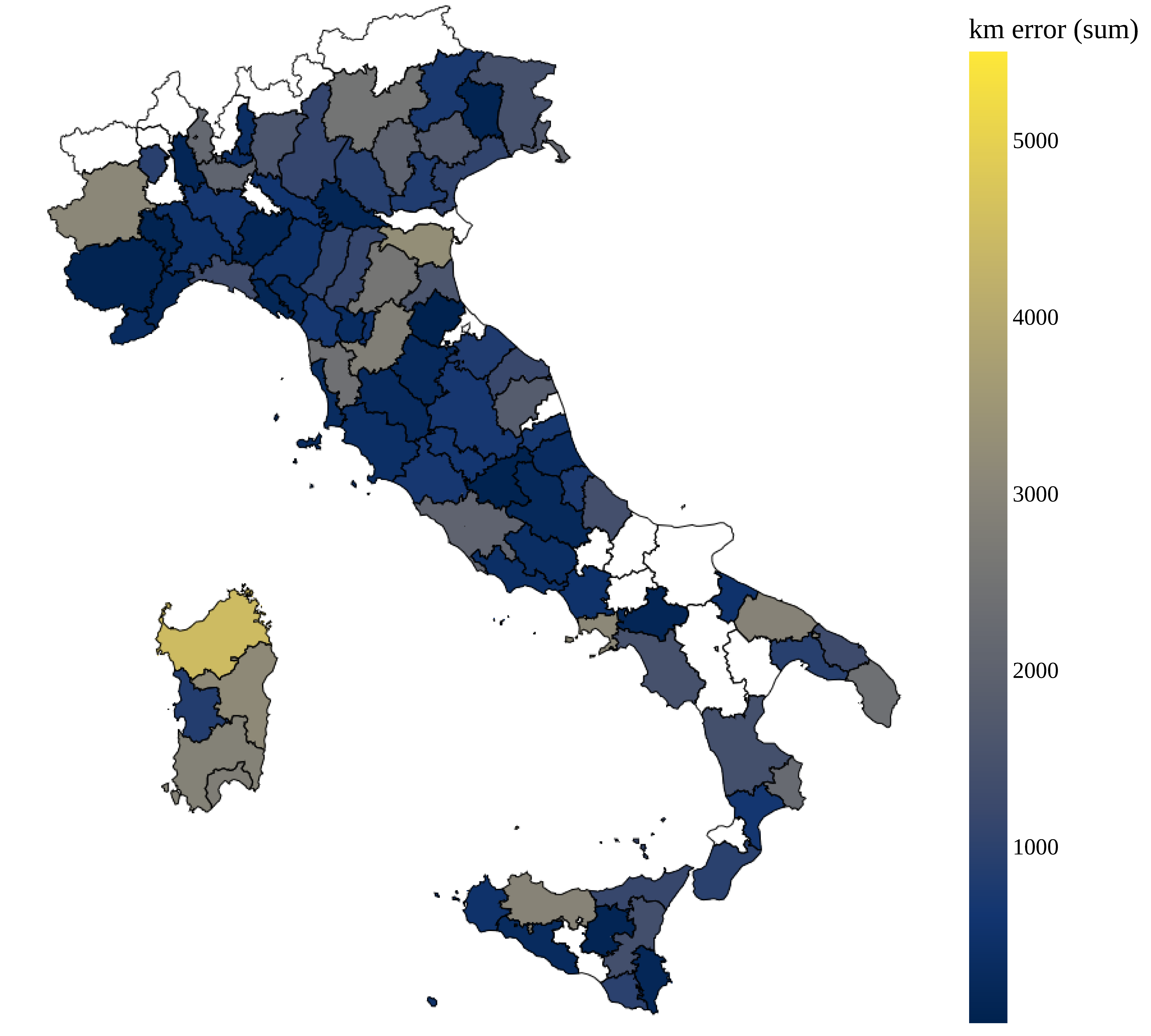}
        \caption{Heathmap of the \textbf{sum} of the regression error (in km) over Italian provinces for Camoscio.}
        \label{fig:regression_error_llama_sum}
     \end{subfigure}
     \hfill
     \begin{subfigure}[t]{.44\linewidth}
        \centering
        \includegraphics[width=\linewidth]{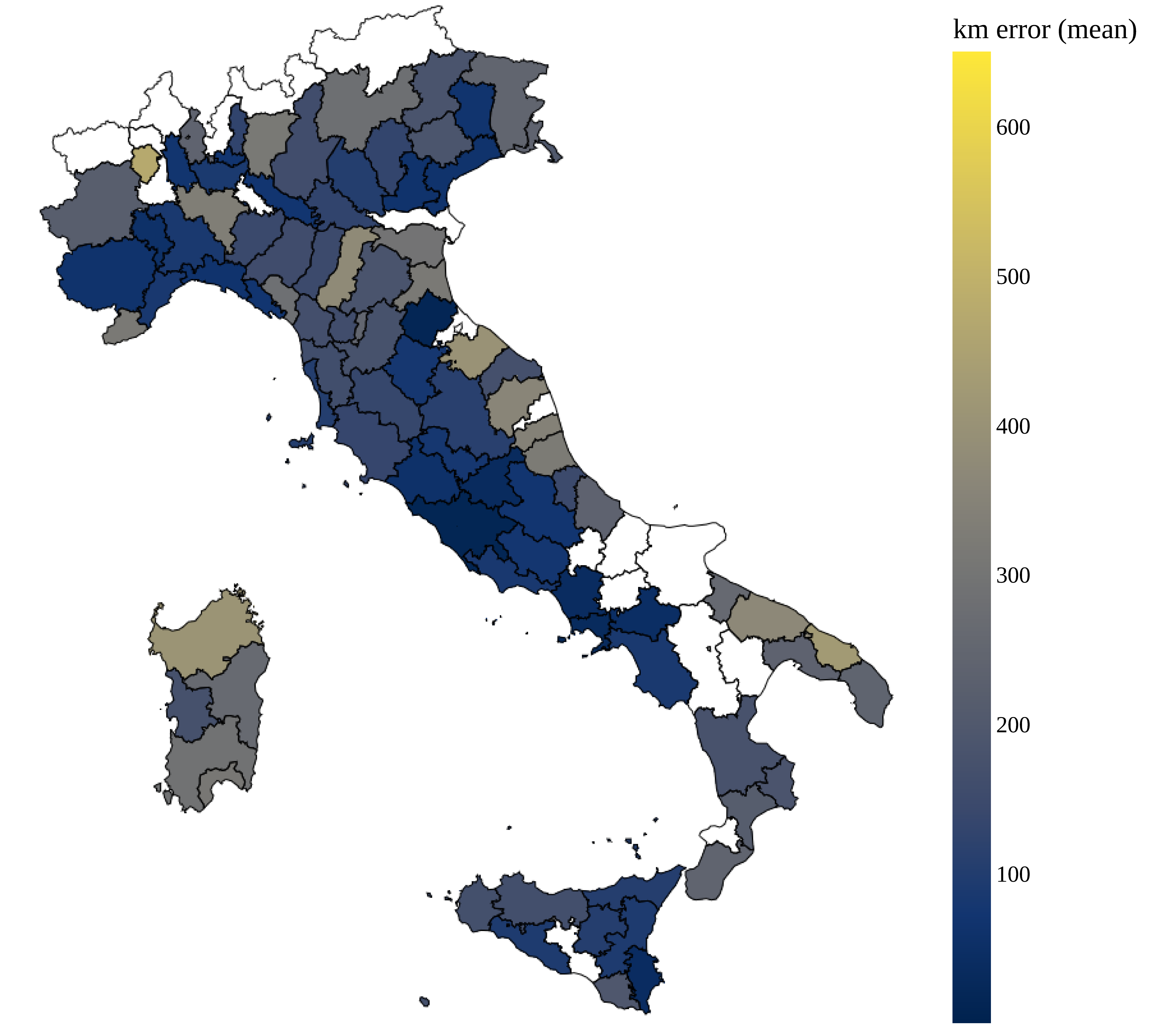}
        \caption{Heathmap of the \textbf{mean} of the regression error (in km) over Italian provinces for Camoscio.}
        \label{fig:regression_error_llama_mean}
     \end{subfigure}
    \vspace{10px}
     \begin{subfigure}[t]{.44\linewidth}     
        \centering
        \includegraphics[width=1\linewidth]{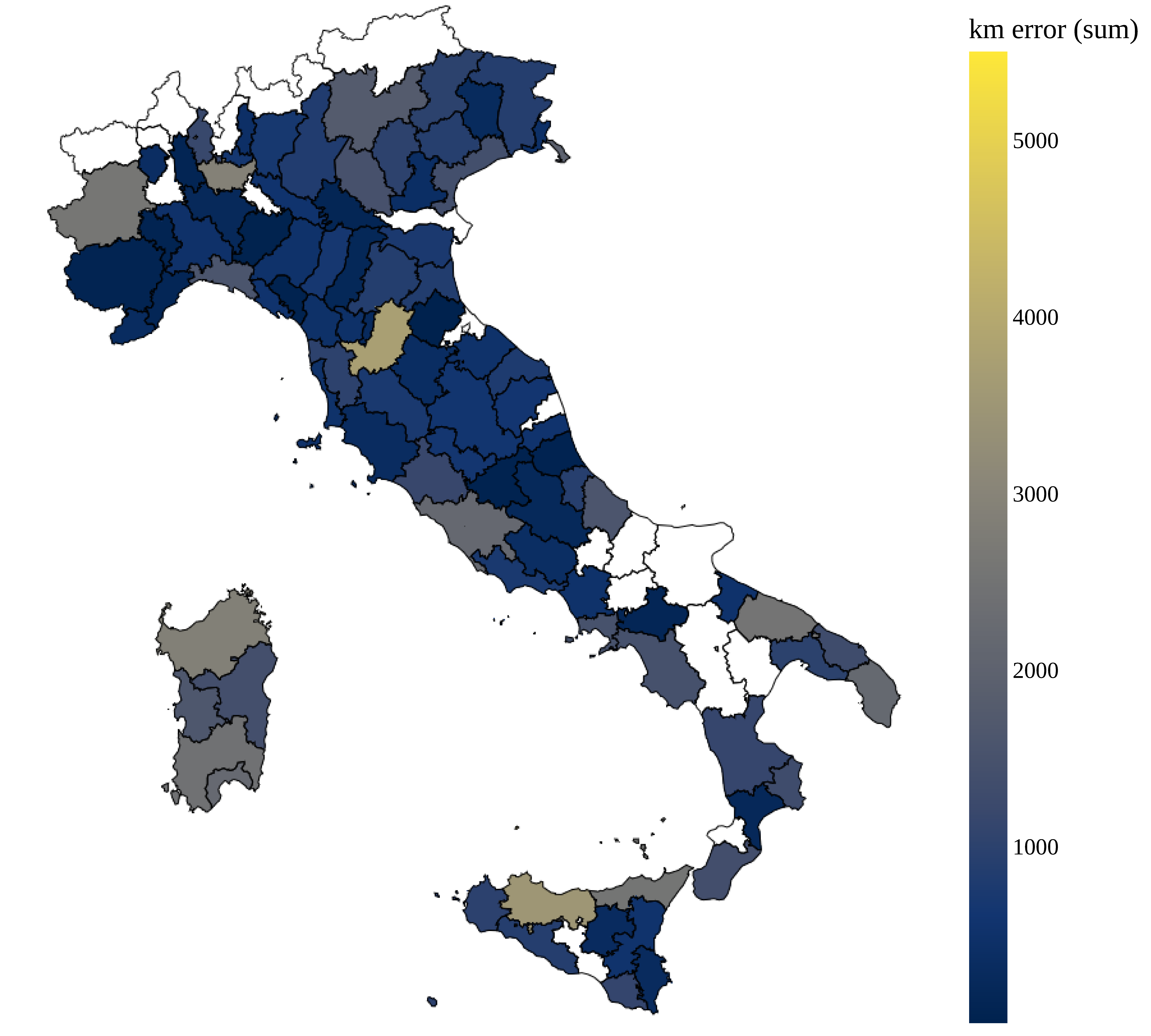}
        \caption{Heathmap of the \textbf{sum} of the regression error (in km) over Italian provinces for ANITA.}
        \label{fig:regression_error_anita_sum}
     \end{subfigure}
     \hfill
     \begin{subfigure}[t]{.44\linewidth}
        \centering
        \includegraphics[width=\linewidth]{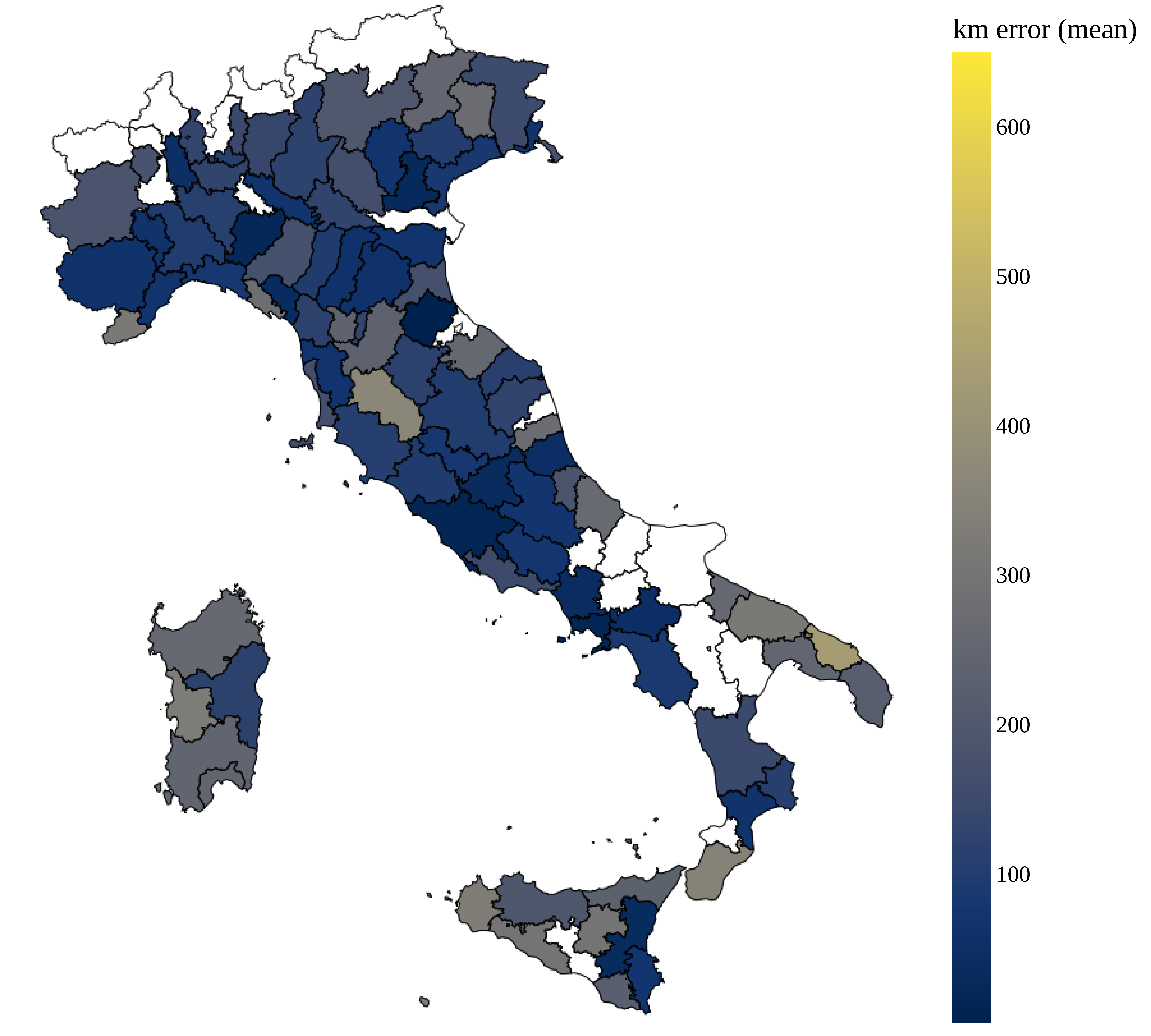}
        \caption{Heathmap of the \textbf{mean} of the regression error (in km) over Italian provinces for ANITA.}
        \label{fig:regression_error_anita_mean}
     \end{subfigure}
     \vspace{10px}
     \begin{subfigure}[t]{.44\linewidth}     
        \centering
        \includegraphics[width=1\linewidth]{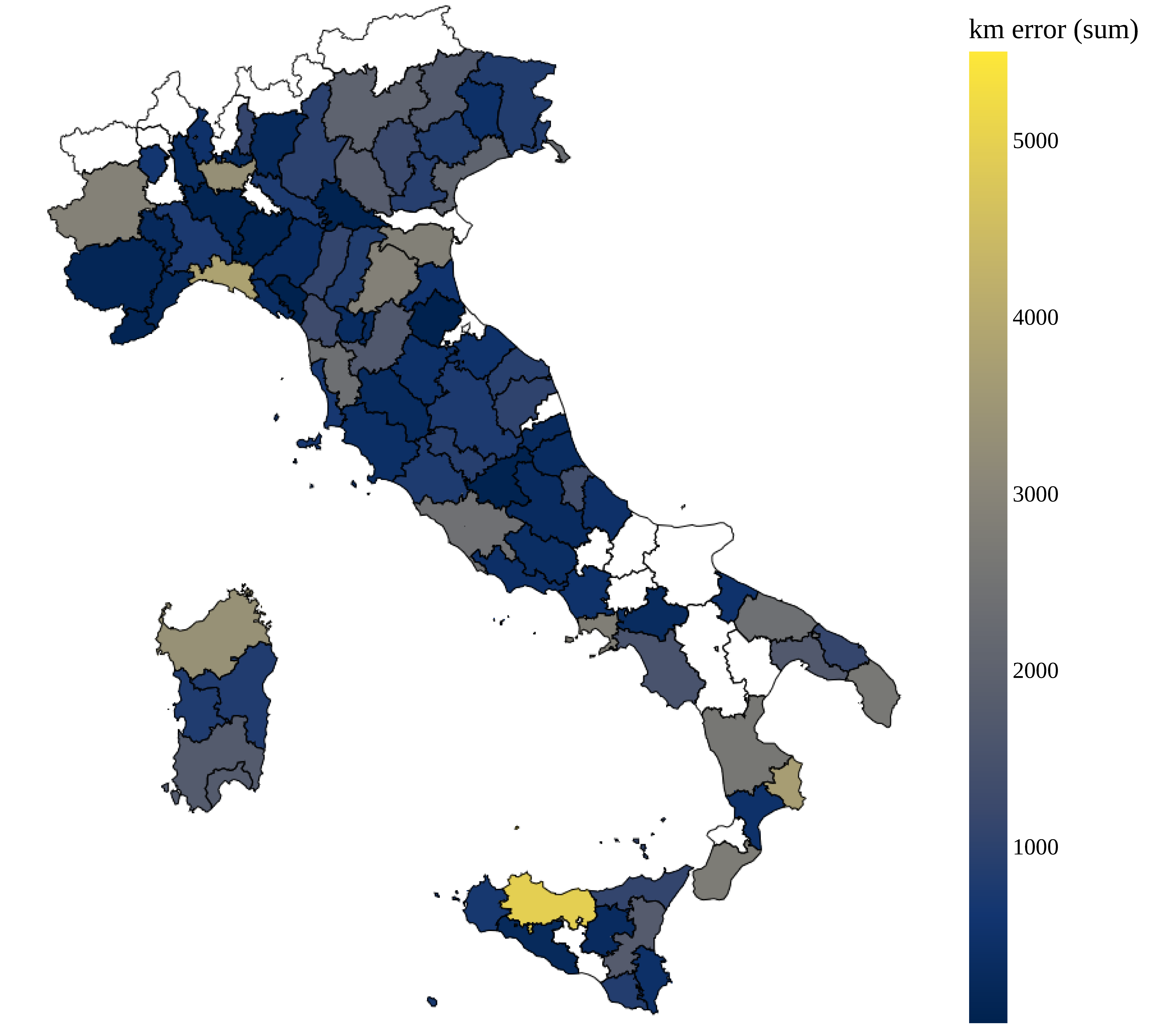}
        \caption{Heathmap of the \textbf{sum} of the regression error (in km) over Italian provinces for Minerva.}
        \label{fig:regression_error_minerva_sum}
     \end{subfigure}
     \hfill
     \begin{subfigure}[t]{.44\linewidth}
        \centering
        \includegraphics[width=\linewidth]{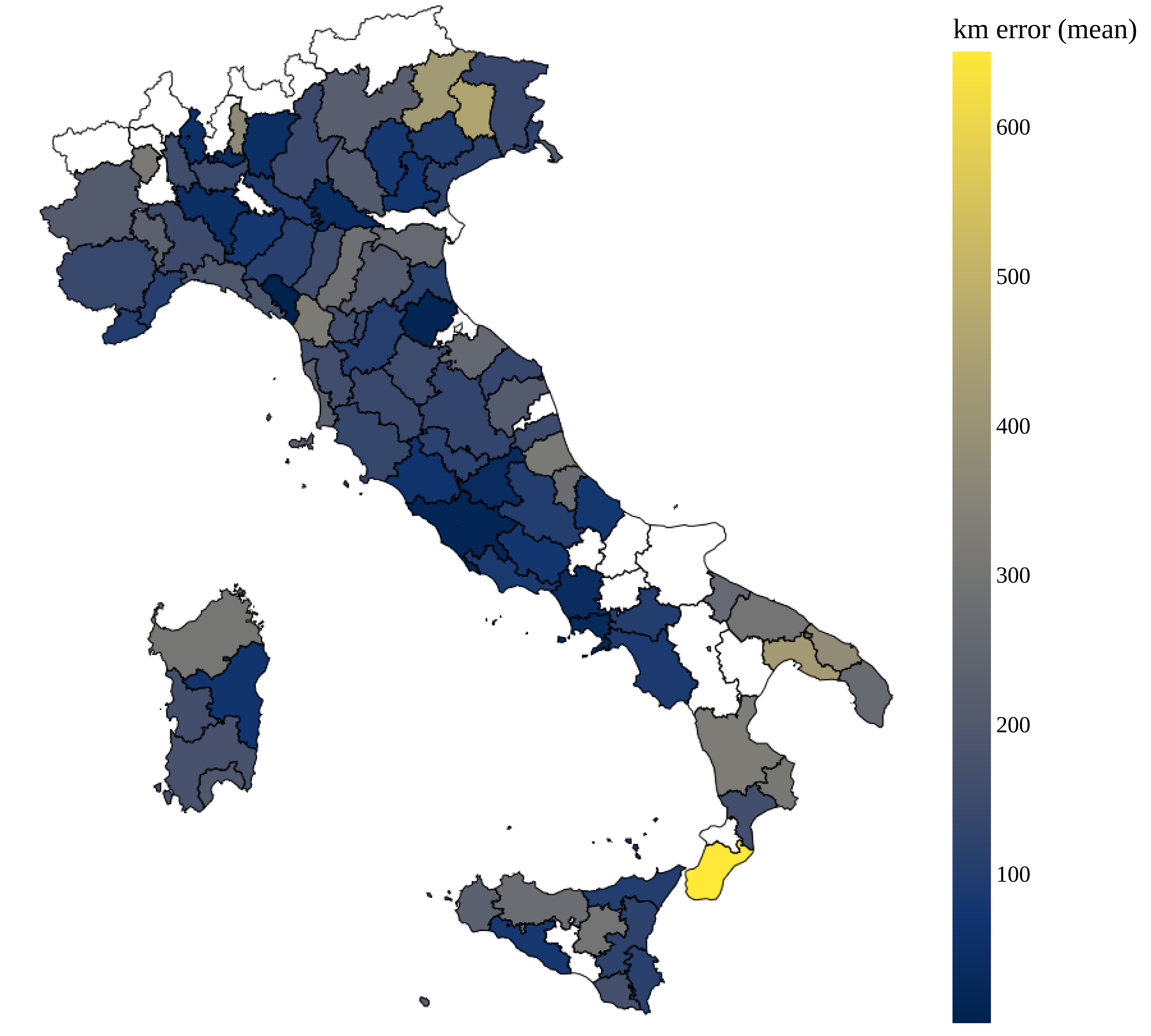}
        \caption{Heathmap of the \textbf{mean} of the regression error (in km) over Italian provinces for Minerva.}
        \label{fig:regression_error_minerva_mean}
     \end{subfigure}
        \caption{Heathmaps of the the regression error (in km) over Italian provinces for all the tested models. The figures in the left column (\ref{fig:regression_error_llama_sum}, \ref{fig:regression_error_anita_sum} and \ref{fig:regression_error_minerva_sum}) show the sum of distance error over the area of a province. Instead, figures in the right column (\ref{fig:regression_error_llama_mean}, \ref{fig:regression_error_anita_mean} and \ref{fig:regression_error_minerva_mean}) show the average distance error over the same areas.}
        \label{fig:regression_error}
\end{figure*}

\end{document}